\documentclass{article}
\usepackage{spconf,graphicx}
\usepackage{multirow}
\usepackage{booktabs}
\usepackage{pifont}
\usepackage{microtype}
\usepackage{amsmath}
\usepackage{amssymb}
\usepackage{url}
\usepackage{enumitem}
\usepackage{xcolor}

\title{PAND: Prompt-Aware Neighborhood Distillation for Lightweight Fine-Grained Visual Classification}
\name{Qiuming Luo$^{\star\dagger}$ \qquad Yuebing Li$^{\star}$ \qquad Feng Li$^{\ddagger}$ \qquad Chang Kong$^{\ddagger\ast}$\thanks{Corresponding author: Chang Kong, E-mail: kongchang@szpu.edu.cn.}}
\address{$^{\star}$Shenzhen University, Shenzhen, China \\
$^{\dagger}$Shenzhen Key Laboratory of Embedded System Design, Shenzhen, China \\
$^{\ddagger}$Shenzhen Polytechnic University, Shenzhen, China}

\begin{document}
%
\maketitle
\begin{abstract}
Distilling knowledge from large Vision-Language Models (VLMs) into lightweight networks is crucial yet challenging in Fine-Grained Visual Classification (FGVC), due to the reliance on fixed prompts and global alignment. To address this, we propose PAND (Prompt-Aware Neighborhood Distillation), a two-stage framework that decouples semantic calibration from structural transfer. First, we incorporate Prompt-Aware Semantic Calibration to generate adaptive semantic anchors. Second, we introduce a neighborhood-aware structural distillation strategy to constrain the student’s local sample-pair logit relation structure. PAND consistently improves over representative KD and VLM-based distillation baselines on four FGVC benchmarks. Notably, our ResNet-18 student achieves 76.09\% accuracy on CUB-200, surpassing the strong baseline VL2Lite by 3.4\%. Code is available at \url{https://github.com/LLLVTA/PAND}.
\end{abstract}

\section{Introduction}\label{sec:intro}
Vision-Language Models (VLMs), such as CLIP~\cite{pmlr-v139-radford21a}, provide strong cross-modal representations for visual recognition. However, their large model size and computational cost hinder deployment in resource-constrained scenarios. In contrast, lightweight architectures (e.g., ResNet and MobileNet) remain practical choices for efficient inference~\cite{He_2016_CVPR,Sandler_2018_CVPR}.

To bridge the performance gap, knowledge distillation (KD) transfers knowledge from high-capacity teachers to lightweight students~\cite{hinton2015distillingknowledgeneuralnetwork,romero2015fitnetshintsdeepnets}. With the development of VLMs, recent works adopt them as teachers by aligning student representations with multimodal semantic spaces~\cite{gu2022,jang2025}. However, existing methods (e.g., VL2Lite~\cite{jang2025}) often underperform in Fine-Grained Visual Classification (FGVC), where subtle visual differences and local sample relations are critical.

We identify two core issues: the semantic gap and the structural gap. These two issues are closely related, as an unadapted semantic space may lead to unreliable local neighborhoods, affecting structural knowledge transfer.

\textbf{Semantic Gap:} Most methods rely on fixed hand-crafted prompts (e.g., ``a photo of a [CLASS]''), which lack adaptability to capture subtle semantic variations among fine-grained categories. Although prompt learning improves semantic representations~\cite{zhou2022e,Khattak_2023_CVPR}, it has not been fully explored in VLM-to-lightweight distillation.

\textbf{Structural Gap:} Existing methods mainly rely on global feature or logit alignment, which may introduce task-irrelevant constraints and fail to preserve local relational structures. Although relational distillation preserves structural information~\cite{park2019,Wu_2023_ICCV}, global modeling remains suboptimal for fine-grained recognition. In FGVC, similar samples often form local neighborhoods where subtle differences determine decision boundaries, making local relation modeling essential.

To address these challenges, we propose PAND (Prompt-Aware Neighborhood Distillation), a two-stage framework that coordinates semantic calibration with structural transfer. In the first stage, Prompt-Aware Semantic Calibration learns task-adaptive prompts while freezing the VLM backbone, producing calibrated semantic anchors. In the second stage, inspired by Neighborhood Logits Relation Distillation (NLRD)~\cite{gou2025}, we develop a neighborhood-aware structural distillation module in the vision-language setting. For each sample, we select its Top-$K$ neighbors based on teacher predictions and align the sample-pair logit relation distributions between teacher and student, enabling effective transfer of fine-grained discrimination ability.

Our main contributions are summarized as follows:
\begin{itemize}[noitemsep, topsep=0pt, parsep=0pt, partopsep=0pt]
    \item We propose PAND, a two-stage framework with Prompt-Aware Semantic Calibration to construct task-adaptive semantic anchors from a frozen VLM, improving semantic alignment for fine-grained distillation.
    \item We develop a sample-level neighborhood-aware structural distillation module that aligns local sample-pair logit relation distributions between teacher and student without modifying the student architecture.
    \item Extensive experiments on four FGVC benchmarks demonstrate that PAND consistently improves over representative KD and VLM-based baselines, achieving up to 3.4\% accuracy improvement over VL2Lite on CUB-200 with ResNet-18.
\end{itemize}

\section{Related Work}\label{sec:related}
\subsection{Vision-Language Model Distillation}
Knowledge Distillation (KD)~\cite{hinton2015distillingknowledgeneuralnetwork} transfers knowledge from a high-capacity teacher to a lightweight student. While early approaches focused on unimodal transfer~\cite{romero2015fitnetshintsdeepnets}, the advent of large-scale Vision-Language Models (VLMs) such as CLIP~\cite{pmlr-v139-radford21a} has shifted the focus toward cross-modal knowledge transfer. Recent works have explored distilling multimodal representations of VLMs into compact vision backbones for efficient open-vocabulary recognition~\cite{gu2022}. Notably, VL2Lite~\cite{jang2025} proposed a task-specific framework to transfer CLIP's knowledge to lightweight networks for image classification by aligning the student's visual features with the image-text embedding space.

However, most existing VLM distillation methods rely on fixed, hand-crafted prompts to construct semantic targets, which limits their ability to capture subtle inter-class variations in FGVC~\cite{zhou2022b}. Furthermore, these methods mainly enforce global feature alignment~\cite{Li_2024_CVPR}, often neglecting local sample-pair logit relation structures. Our work addresses these limitations by incorporating task-adaptive prompt learning and neighborhood-aware structural constraints.

\subsection{Prompt Learning for VLMs}
CoOp (Context Optimization)~\cite{zhou2022b} introduced learnable continuous context vectors to replace manual text prompts, significantly improving CLIP's adaptation to downstream tasks. Subsequent variants such as CoCoOp~\cite{zhou2022e} further enhanced generalization by conditioning prompts on image instances. These methods primarily utilize prompt learning to improve VLM inference performance. In contrast, our PAND framework leverages prompt learning as a semantic anchor calibration mechanism during teacher preparation. By freezing the VLM backbone and optimizing only the prompts, we generate task-specific semantic representations that provide precise supervision for student distillation.

\subsection{Relational and Neighborhood Knowledge Distillation}
Beyond logit-based~\cite{hinton2015distillingknowledgeneuralnetwork} and feature-based~\cite{romero2015fitnetshintsdeepnets} distillation, Relational Knowledge Distillation (RKD)~\cite{park2019} transfers structural relations among data samples. However, conventional RKD often models global relationships, which may be challenging for lightweight students~\cite{Chen_2021_CVPR}. Recent studies emphasize preserving local neighborhood structures. For example, NRKD~\cite{gou2025} and Local Correlation Distillation~\cite{li2020} model relations between samples and their nearest neighbors. Building on prior neighborhood relation distillation methods such as NLRD~\cite{gou2025}, we introduce a neighborhood-aware structural distillation module for vision-language distillation under prompt-calibrated semantic supervision. Our method aligns the student's local sample-pair logit relation with that of the VLM teacher under calibrated semantic supervision, enabling more effective transfer of fine-grained discrimination ability.

\section{Methodology}\label{sec:method}
\begin{figure}[t!] 
    \centering 
    \includegraphics[width=1.0\linewidth]{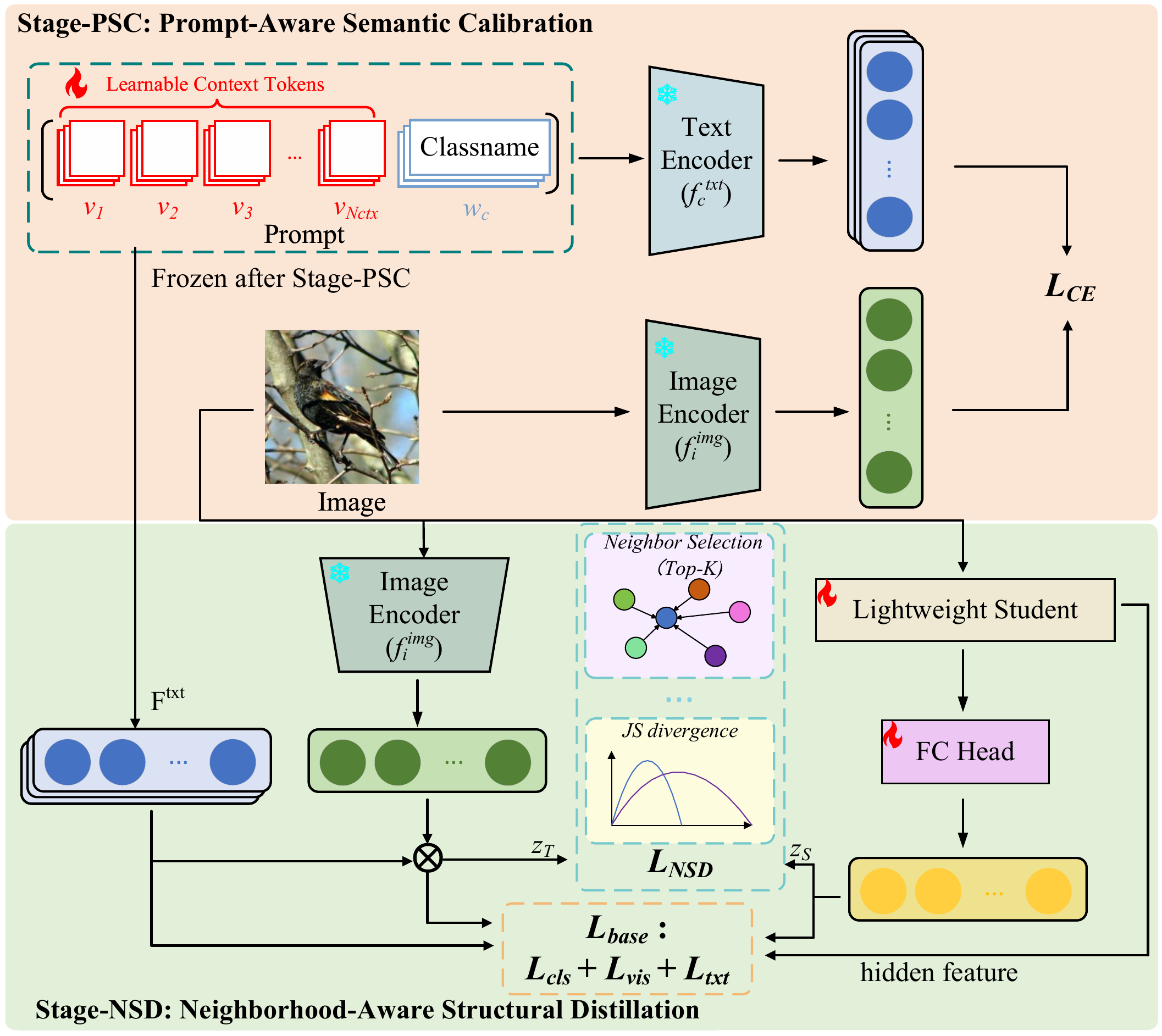}
    \vspace{-15pt}
    \caption{\textbf{The overall framework of PAND.} The training is decoupled into two stages. \textbf{Stage-PSC:} We learn task-specific context tokens to generate calibrated text features (semantic anchors) while keeping the VLM encoders frozen. \textbf{Stage-NSD:} Using the learned text features as a fixed classifier for the teacher, we train the lightweight student. The student is supervised by the VL2Lite base loss~\cite{jang2025} and our proposed Neighborhood-Aware Structural Distillation, which aligns the student's local sample-pair logit relation distribution with that induced by the teacher.} 
    \label{fig:framework}
    \vspace{-15pt}
\end{figure}

In this section, we present PAND, a two-stage distillation framework designed to transfer multimodal knowledge from large-scale Vision-Language Models (VLMs) to lightweight student networks for fine-grained visual classification. The overall architecture is illustrated in Fig.~\ref{fig:framework}.

\subsection{Overview}
Existing VLM distillation methods rely on fixed prompts and global alignment, which are suboptimal for FGVC. We therefore adopt a two-stage framework:

\textbf{Stage-PSC (Prompt-Aware Semantic Calibration).}
We employ Context Optimization (CoOp)~\cite{zhou2022b} to learn task-adaptive semantic anchors while keeping the VLM encoders frozen. This yields a set of stable and discriminative text features tailored to the target dataset.

\textbf{Stage-NSD (Neighborhood-Aware Structural Distillation).}
We freeze the learned semantic anchors and the teacher model to supervise the lightweight student. In addition to standard feature alignment (as in VL2Lite~\cite{jang2025}), we introduce a neighborhood-aware structural distillation module to explicitly constrain the student's local sample-pair logit relation structure, encouraging it to preserve the teacher-induced relation structure over neighboring samples.

\subsection{Stage-PSC: Prompt-Aware Semantic Calibration}
In the first stage, our goal is to construct a semantic space that is more discriminative than one derived from generic hand-crafted prompts. We adopt the CoOp paradigm~\cite{zhou2022b} to optimize continuous context tokens. Formally, for a specific class $c$, the prompt is parameterized as a sequence of $N_{ctx}$ learnable context vectors $\{\mathbf{v}_1, \mathbf{v}_2, \ldots, \mathbf{v}_{N_{\text{ctx}}}\}$ followed by the fixed class name embedding:
\begin{equation}
\mathbf{p}_c = [\mathbf{v}_1, \mathbf{v}_2, \ldots, \mathbf{v}_{N_{\text{ctx}}}, \mathbf{w}_c],
\end{equation}
where $\mathbf{w}_c$ represents the embedding of the $c$-th class name.

During training, we utilize a pre-trained VLM (e.g., CLIP~\cite{pmlr-v139-radford21a}) consisting of an image encoder $E_{img}$ and a text encoder $E_{txt}$. Both encoders are frozen to preserve the pre-trained multimodal knowledge. For an input image $x_i$, the image encoder extracts the visual feature $\mathbf{f}^{img}_i = E_{img}(x_i)$. Simultaneously, the text encoder maps the learnable prompt $\mathbf{p}_c$ to the text feature $\mathbf{f}^{txt}_c = E_{txt}(\mathbf{p}_c)$. All features are $\ell_2$-normalized.

The optimization objective is to maximize the similarity between the image feature and the correct class text feature using an image-to-text cross-entropy loss (in the spirit of contrastive VLM training~\cite{pmlr-v139-radford21a,pmlr-v139-jia21b}):
\begin{equation}
\mathcal{L}_{\text{CE}}=
-\frac{1}{N}\sum_{i=1}^{N}
\log
\frac{
\exp\!\left(\langle \mathbf{f}^{img}_{i},\mathbf{f}^{txt}_{y_i}\rangle/\tau\right)
}{
\sum_{c=1}^{C}
\exp\!\left(\langle \mathbf{f}^{img}_{i},\mathbf{f}^{txt}_{c}\rangle/\tau\right)
},
\end{equation}

After Stage-PSC, we obtain a set of optimized text features $\mathbf{F}^{txt} = [\mathbf{f}_1^{txt}, \ldots, \mathbf{f}_C^{txt}]$, which serve as fixed semantic anchors for the subsequent stage.

\subsection{Stage-NSD: Neighborhood-Aware Structural Distillation}
In Stage-NSD, we train a lightweight student network using the frozen VLM teacher and the calibrated semantic anchors obtained from Stage-PSC. 
The objective is to transfer not only global semantic supervision, but also the teacher-induced local sample-pair logit relation structure to the student.

\paragraph*{Teacher and Student Architectures.}
The teacher consists of the frozen VLM image encoder, e.g., CLIP~\cite{pmlr-v139-radford21a}, and the fixed text features $\mathbf{F}^{txt}$. 
For an input image $x_i$, the teacher produces a normalized visual feature $\mathbf{f}^{img}_i$ and generates logits by projecting it onto the prompt-calibrated text anchors:
\begin{equation}
\mathbf{z}^{(i)}_{T}
=
\mathbf{f}^{img}_{i}(\mathbf{F}^{txt})^{\top},
\end{equation}
Since both image and text features are $\ell_2$-normalized, the teacher logits correspond to cosine similarities between the image feature and the prompt-calibrated text anchors.

The student model comprises a lightweight visual backbone $S_{\text{img}}$ and a fully connected classification head:
\begin{equation}
\mathbf{z}^{(i)}_{S}
=
\mathrm{FC}(\mathbf{f}^{S}_{i}),
\end{equation}

\paragraph*{Global Alignment Loss.}
Following VL2Lite~\cite{jang2025}, we apply a combined loss to align the student's representation with the teacher globally:
\begin{equation}
\mathcal{L}_{\text{base}}
=
\lambda_{\text{cls}}\mathcal{L}_{\text{cls}}
+
\lambda_{\text{vis}}\mathcal{L}_{\text{vis}}
+
\lambda_{\text{txt}}\mathcal{L}_{\text{txt}},
\end{equation}
Here, $\mathcal{L}_{\text{cls}}$ denotes the standard classification loss, while $\mathcal{L}_{\text{vis}}$ and $\mathcal{L}_{\text{txt}}$ follow the visual and textual alignment losses in VL2Lite~\cite{jang2025}.

\paragraph*{Neighborhood-Aware Structural Distillation.}

Given a mini-batch $\mathcal{B}=\{(x_i,y_i)\}_{i=1}^{B}$, we first compute the teacher logits $\mathbf{z}^{(i)}_{T}$ and student logits $\mathbf{z}^{(i)}_{S}$ for each sample. 
For a query sample $x_i$, we construct its neighborhood set $\mathcal{N}_i$ by selecting the Top-$K$ most similar samples under the teacher logit space, excluding the query sample itself:
\begin{equation}
\mathcal{N}_i
=
\operatorname{TopK}_{x_j \in \mathcal{B},\, j\neq i}
\left(
\mathrm{sim}
\left(
\mathbf{z}^{(i)}_{T},
\mathbf{z}^{(j)}_{T}
\right)
\right),
\end{equation}
where $\mathrm{sim}(\cdot,\cdot)$ denotes cosine similarity. The query sample itself is excluded to avoid the trivial self-match. 
The neighborhood is selected only from the teacher logit space, so the same teacher-selected neighbors are used to construct both teacher and student relation distributions.

Given the same neighborhood set $\mathcal{N}_i$, we construct a sample-pair logit relation distribution between the query sample $x_i$ and each neighboring sample $x_j$. 
Inspired by neighborhood logits relation distillation~\cite{gou2025}, we characterize the relation between two samples by the difference between their logit vectors, rather than by a single scalar similarity value:
\begin{equation}
\begin{aligned}
\boldsymbol{\rho}^{T}_{ij}
&=\operatorname{softmax}\!\left(
\frac{\mathbf{z}^{(i)}_{T}-\mathbf{z}^{(j)}_{T}}{\tau_r}
\right),\\
\boldsymbol{\rho}^{S}_{ij}
&=\operatorname{softmax}\!\left(
\frac{\mathbf{z}^{(i)}_{S}-\mathbf{z}^{(j)}_{S}}{\tau_r}
\right),
\end{aligned}
\end{equation}

where $x_j \in \mathcal{N}_i$, $\tau_r$ is the temperature, and $\boldsymbol{\rho}^{T}_{ij}$ and $\boldsymbol{\rho}^{S}_{ij}$ denote relation distributions.

We align these two sample-pair relation distributions using the Jensen--Shannon divergence:
\begin{equation}
\mathcal{L}_{\text{NSD}}
=
\frac{1}{BK}
\sum_{i=1}^{B}
\sum_{x_j \in \mathcal{N}_i}
\mathrm{JS}
\left(
\boldsymbol{\rho}^{T}_{ij}
\parallel
\boldsymbol{\rho}^{S}_{ij}
\right),
\end{equation}

This loss encourages the student to preserve the teacher-induced local sample-pair logit relation in the calibrated vision-language prediction space.

\subsection{Overall Optimization Objective}
The final training objective for Stage-NSD is given by:
\vspace{-4pt}
\begin{equation}
\mathcal{L}_{\text{total}}
=
\mathcal{L}_{\text{base}}
+
\lambda_{\text{NSD}}
\mathcal{L}_{\text{NSD}},
\end{equation}
\vspace{-4pt}
where $\lambda_{\text{NSD}}$ is the hyper-parameter to balance two losses.

\subsection{Analysis of the Two-Stage Strategy}
We adopt a decoupled two-stage strategy to ensure optimization stability. In Stage-PSC, prompt learning requires clean gradients from the pre-trained VLM~\cite{pmlr-v139-radford21a} to converge to accurate semantic anchors, as commonly observed in prompt tuning paradigms~\cite{zhou2022b,zhou2022e}. Joint optimization with Stage-NSD would introduce noisy gradients from the randomly initialized student and the distillation losses, destabilizing prompt learning. By freezing the semantic anchors after Stage-PSC, we provide a stationary and high-quality target for the student, enabling robust convergence in Stage-NSD. 

\section{Experiments}\label{sec:exp}
\subsection{Experimental Setup}
\textbf{Datasets.} We evaluate PAND on four challenging fine-grained visual classification (FGVC) benchmarks: CUB-200-2011 (200 bird species)~\cite{wah_branson_welinder_perona_belongie_2011}, Oxford-IIIT Pet (37 cat and dog breeds)~\cite{parkhi2012cats}, Stanford Dogs (120 dog breeds)~\cite{khosla2011novel}, and FGVC-Aircraft (100 aircraft variants)~\cite{maji2013fine}. We utilize the official training and testing splits for all datasets.

\textbf{Architectures.}
We use CLIP ConvNeXt-XXL~\cite{liu2022convnet} pre-trained on \texttt{laion2b\allowbreak\_s34b\allowbreak\_b82k\allowbreak\_augreg\allowbreak\_soup} as the frozen teacher, with task-specific prompts learned in Stage-PSC. ResNet-18 and MobileNet-V2, initialized with ImageNet-1k pre-trained weights~\cite{deng2009imagenet}, are used as lightweight students.

\textbf{Implementation Details.}
Our framework is implemented in PyTorch on four NVIDIA V100 GPUs. In Stage-PSC, the VLM encoders are frozen and only the context tokens ($N_{ctx}=16$) are optimized using CoOp~\cite{zhou2022b} with SGD (lr=0.002, momentum=0.9, weight decay=0) for 200 epochs and batch size 128. In Stage-NSD, the teacher and semantic anchors are frozen, and the student is trained for 300 epochs using AdamW~\cite{loshchilov2018decoupled} (lr=$1\times10^{-4}$, weight decay=$1\times10^{-4}$) with cosine annealing~\cite{loshchilov2016sgdr} (min lr=$1\times10^{-5}$). We set $\tau=2.0$, $K=3$, $\tau_r=1.0$, and $\lambda_{\text{NSD}}=1.0$. Following VL2Lite~\cite{jang2025}, $\lambda_{cls}=0.01$ and $\lambda_{vis}=\lambda_{txt}=0.495$.

\subsection{Comparison with Representative Baselines}
We compare PAND with the baseline (training without KD), standard KD~\cite{hinton2015distillingknowledgeneuralnetwork}, Relational KD (RKD)~\cite{park2019}, as well as several representative VLM-based distillation methods, including VL2Lite~\cite{jang2025} and RISE~\cite{Huang_2023_ICCV}. Table~\ref{tab:comp_sota} presents the Top-1 classification accuracy across four fine-grained benchmarks.

\textbf{Overall Performance.}  
As shown in Table~\ref{tab:comp_sota}, PAND achieves the best performance across all datasets and student architectures. Compared with conventional distillation methods, our approach yields consistent improvements, demonstrating the effectiveness of prompt-based semantic calibration and neighborhood-aware structural supervision.

\textbf{ResNet-18.}  
On CUB-200, PAND achieves 76.09\%, surpassing the w/o KD baseline by 11.61\% and VL2Lite by 3.42\%. Notably, on the nearly saturated Oxford Pets dataset, PAND still improves upon VL2Lite by about 0.4\%.

\textbf{MobileNet-V2.}  
PAND maintains strong performance on compact models. On CUB-200, it achieves 76.52\%, outperforming VL2Lite by 4.33\%. Moreover, on FGVC-Aircraft, PAND exceeds VL2Lite by 5.7\%, highlighting its advantage for compact student models. 

Overall, these results demonstrate that PAND effectively enhances the performance of lightweight models for fine-grained recognition, particularly in scenarios requiring precise semantic understanding and robust local structural modeling.

\begin{table}[t]
    \centering
    \caption{Top-1 Accuracy (\%) comparison on four fine-grained benchmarks. The best results are highlighted in \textbf{bold}.}
    \label{tab:comp_sota}
    \resizebox{\linewidth}{!}{
    \begin{tabular}{l|l|cccc}
    \hline
    \textbf{Student} & \textbf{Method} & \textbf{CUB-200} & \textbf{Oxford Pets} & \textbf{Aircraft} & \textbf{Dogs} \\ \hline
    \multirow{6}{*}{ResNet-18}
    & w/o KD & 64.48 & 84.94 & 55.21 & 67.37 \\
    & KD & 70.95 & 86.74 & 53.83 & 68.80 \\
    & RKD & 68.31 & 86.94 & 50.98 & 69.03 \\
    & RISE & 69.69 & 86.81 & 54.81 & 68.72 \\
    & VL2Lite & 72.67 & 88.56 & 60.82 & 73.14 \\ \cline{2-6}
    & \textbf{PAND (Ours)} & \textbf{76.09} & \textbf{88.97} & \textbf{63.25} & \textbf{74.98} \\ \hline
    
    \multirow{6}{*}{MobileNet-V2}
    & w/o KD & 65.42 & 84.07 & 53.67 & 68.28 \\
    & KD & 68.00 & 86.07 & 48.78 & 73.02 \\
    & RKD & 67.95 & 86.56 & 50.29 & 69.45 \\
    & RISE & 67.51 & 86.03 & 52.44 & 69.28 \\
    & VL2Lite & 72.19 & 87.55 & 58.99 & 73.28 \\ \cline{2-6}
    & \textbf{PAND (Ours)} & \textbf{76.52} & \textbf{88.28} & \textbf{64.75} & \textbf{74.52} \\ \hline
        \end{tabular}
    }
    \vspace{-15pt}
\end{table}

\subsection{Ablation Study}
To verify the contribution of each component in our proposed PAND framework, we conducted ablation studies on the CUB-200-2011 dataset using ResNet-18 as the student network. All other experimental settings follow the main experiments. The results are summarized in Table~\ref{tab:ablation}.

\textbf{Effect of Prompt-Aware Semantic Calibration (Stage-PSC).}
As shown in Table~\ref{tab:ablation}, introducing task-adaptive prompt learning into the VL2Lite baseline improves the accuracy from 72.67\% to 73.52\%. Compared to fixed hand-crafted prompts, learnable context tokens generate more discriminative semantic anchors, providing more accurate supervision for fine-grained categories~\cite{zhou2022b}.

\textbf{Effect of Neighborhood-Aware Structural Distillation (Stage-NSD).}
When introducing the NSD module alone (without Stage-PSC), the performance increases to 75.91\%. This improvement is more substantial than using prompt learning alone, indicating that modeling local sample-pair logit relations is beneficial~\cite{gou2025,li2020}. It encourages the student to preserve the teacher-induced sample-pair relation structure over neighboring samples.

\textbf{Combined Effect of Two Stages.}
When both PSC and NSD are employed, the full PAND framework achieves the best accuracy of 76.09\%. Compared with using NSD alone, PSC provides an additional improvement, suggesting that task-adaptive semantic anchors can further benefit neighborhood-aware sample-pair relation transfer. 

\begin{table}[h]
    \centering
    \vspace{-10pt}
    \caption{Ablation study of different components on the CUB-200 dataset with ResNet-18 student. \textbf{PSC}: Prompt-Aware Semantic Calibration; \textbf{NSD}: Neighborhood-Aware Structural Distillation.}
    \label{tab:ablation}
    \begin{tabular}{l|cc|c}
    \toprule
    \textbf{Method} & \textbf{PSC} & \textbf{NSD} & \textbf{Accuracy (\%)} \\
    \midrule
    VL2Lite (Baseline) & \ding{55} & \ding{55} & 72.67 \\
    Baseline + PSC & \ding{51} & \ding{55} & 73.52 \\
    Baseline + NSD & \ding{55} & \ding{51} & 75.91 \\
    \textbf{PAND (Ours)} & \ding{51} & \ding{51} & \textbf{76.09} \\
    \bottomrule
    \end{tabular}
    \vspace{-10pt}
\end{table}

\subsection{Sensitivity Analysis of Structural Loss Weight}

\begin{figure}[t]
    \centering
    \includegraphics[width=0.9\linewidth]{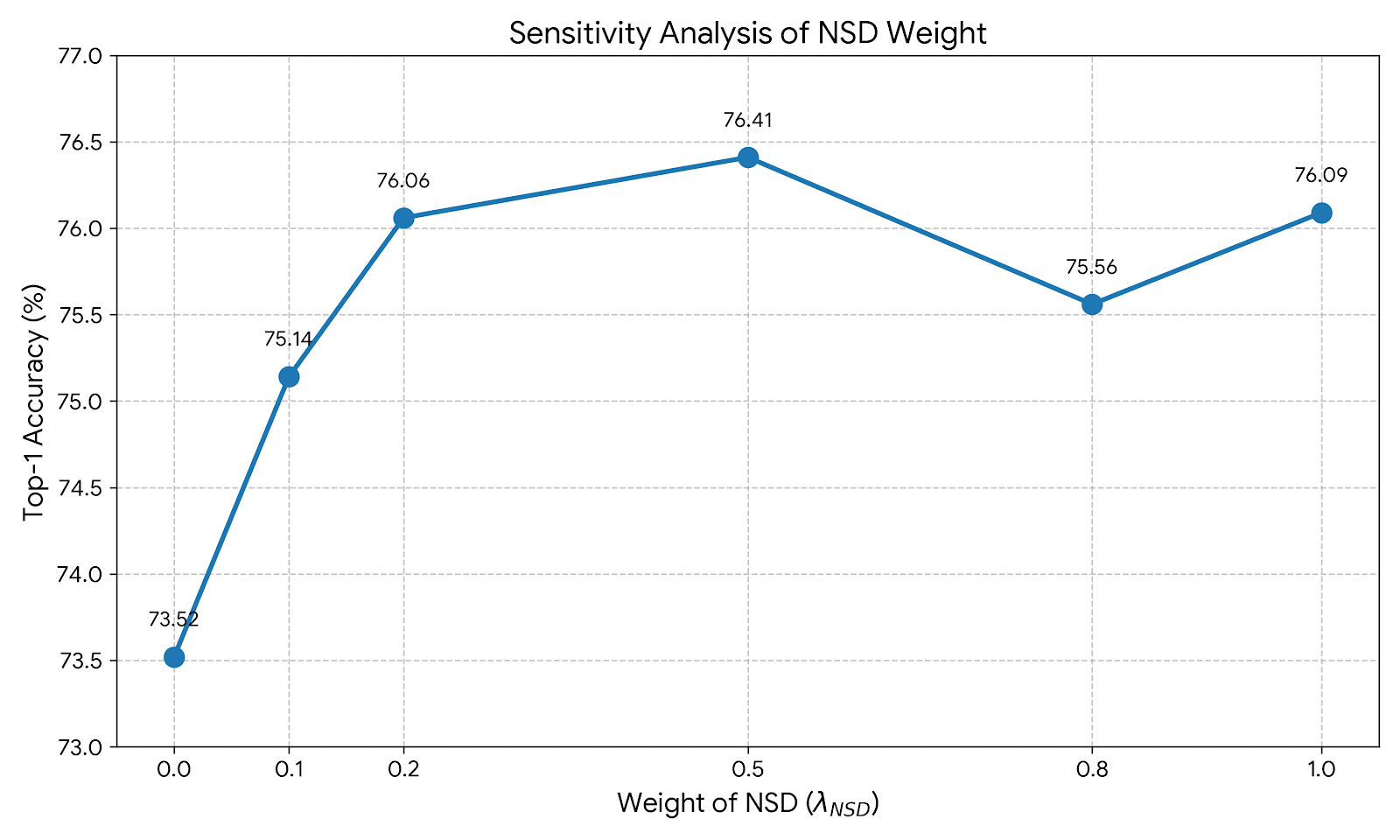}
    \vspace{-15pt}
    \caption{Sensitivity analysis of the NSD weight $\lambda_{\text{NSD}}$ on CUB-200 with ResNet-18.}
    \label{fig:sensitivity}
    \vspace{-15pt}
\end{figure}

We analyze the sensitivity of PAND to the structural loss weight $\lambda_{\text{NSD}}$ on CUB-200 with ResNet-18. As shown in Fig.~\ref{fig:sensitivity}, removing the structural loss ($\lambda_{\text{NSD}}=0$) leads to inferior performance. As $\lambda_{\text{NSD}}$ increases from 0 to 0.5, the accuracy consistently improves, reaching the best 76.41\% at $\lambda_{\text{NSD}}=0.5$. Further increasing $\lambda_{\text{NSD}}$ does not consistently improve performance, suggesting that overly strong structural supervision may be suboptimal. 

\subsection{Analysis and Visualization}

\begin{figure}[t]
  \centering 
  \includegraphics[width=\linewidth]{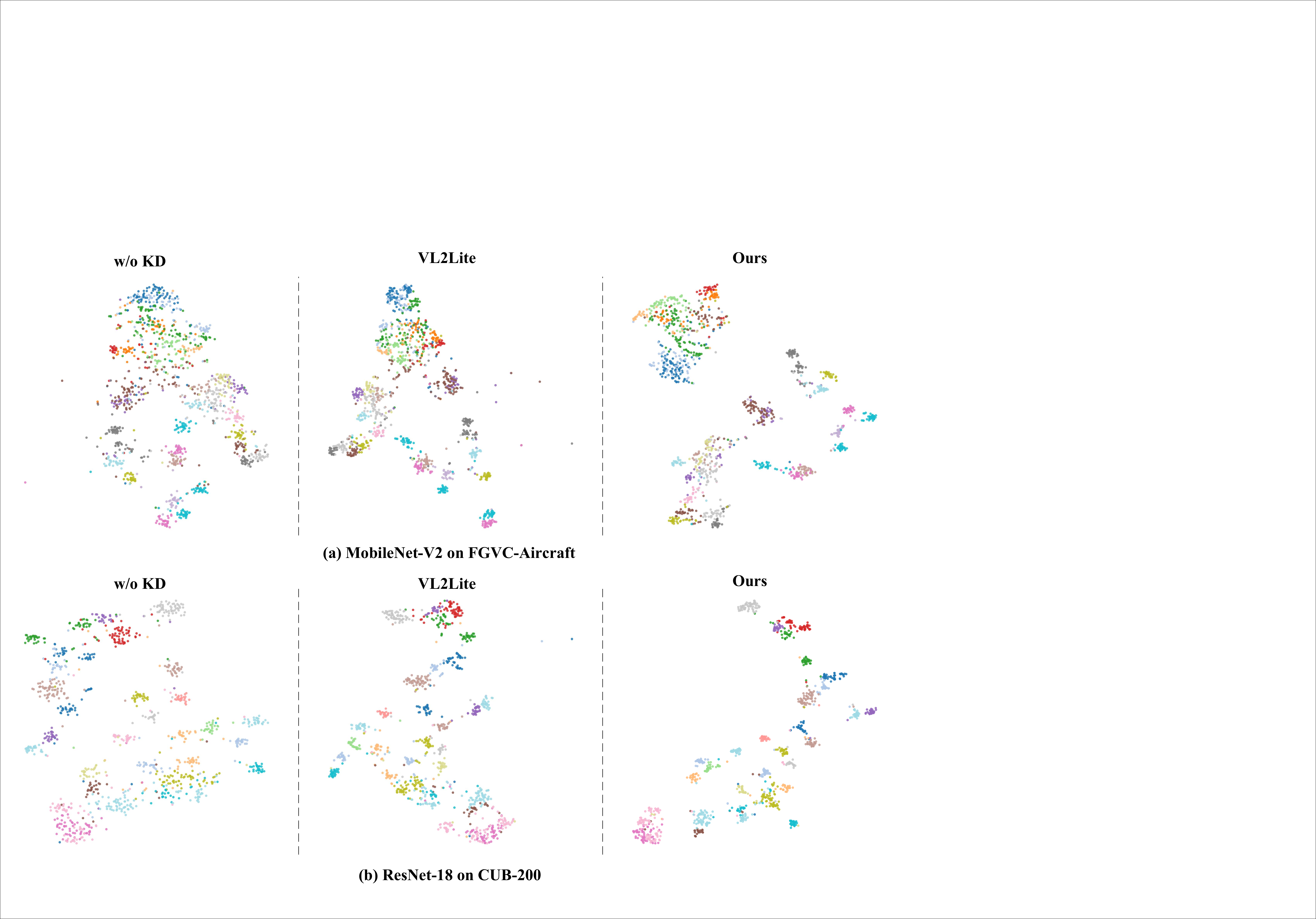}
  \vspace{-15pt}
  \caption{t-SNE visualization of feature distributions. (a) MobileNet-V2 on FGVC-Aircraft. (b) ResNet-18 on CUB-200. Each subplot compares w/o KD, VL2Lite, and our method. Different colors indicate different categories.}
  \label{fig:tsne}
  \vspace{-15pt}
\end{figure}

We visualize the learned feature embeddings using t-SNE~\cite{vandermaaten2008visualizing} on MobileNet-V2 (FGVC-Aircraft) and ResNet-18 (CUB-200), as shown in Fig.~\ref{fig:tsne}.

\textbf{Baseline (w/o KD).} Models trained without distillation exhibit scattered feature distributions, with weak intra-class compactness and substantial overlap among similar categories, indicating limited discriminative capability.

\textbf{VL2Lite Distillation.} With VL2Lite~\cite{jang2025}, the feature space becomes more structured, and several categories form clearer clusters. However, visually similar classes remain partially entangled, suggesting that global alignment alone is insufficient.

\textbf{PAND Distillation (Ours).} In contrast, PAND produces more compact and well-separated clusters across both datasets. Samples from the same class are tightly grouped, while clear margins emerge between different classes, demonstrating improved fine-grained discrimination.

\section{Conclusion}\label{sec:conclusion}
In this paper, we propose \textbf{PAND}, a two-stage distillation framework for transferring multimodal knowledge from large-scale vision-language models to lightweight networks for fine-grained visual classification. PAND decouples semantic calibration from structural transfer by addressing fixed-prompt and global-only alignment limitations: Prompt-Aware Semantic Calibration generates task-adaptive semantic anchors with a frozen VLM backbone, while a neighborhood-aware structural module aligns local sample-pair logit relations between teacher and student. Experiments on four FGVC benchmarks show that PAND consistently improves over representative KD and VLM-based baselines, enabling ResNet-18 and MobileNet-V2 to achieve notable accuracy gains.

\section{Acknowledgements}
This work was supported by the Shenzhen Key Laboratory of Embedded System Design, the Shenzhen Key Laboratory of Service Computing and Applications, the Post-doctoral Later-stage Foundation Project of Shenzhen Polytechnic University (Grant No. 6023271039K).

\bibliographystyle{IEEEbib}
\bibliography{strings,refs}

@Article{zhou2022b,
  author = 	 "Zhou, Kaiyang and Yang, Jingkang and Loy, Chen Change and others",
  title = 	 "Learning to Prompt for Vision-Language Models",
  journal = 	 "International Journal of Computer Vision",
  year = 	 "2022",
  volume = 	 "130",
  number = 	 "9",
  pages = 	 "2337--2348"
}

@InProceedings{zhou2022e,
  author = 	 "Zhou, Kaiyang and Yang, Jingkang and Loy, Chen Change and Liu, Ziwei",
  title = 	 "Conditional Prompt Learning for Vision-Language Models",
  booktitle = 	 "Proc. CVPR",
  year = 	 "2022",
  pages = 	 "16816-16825"
}

@inproceedings{jang2025,
  title = {{VL2Lite}: Task-Specific Knowledge Distillation from Large Vision-Language Models to Lightweight Networks},
  booktitle = {Proc. CVPR},
  author = {Jang, Jinseong and Ma, Chunfei and Lee, Byeongwon},
  year = 2025,
  pages = {30073--30083}
}

@inproceedings{park2019,
  title = {Relational Knowledge Distillation},
  booktitle = {Proc. CVPR},
  author = {Park, Wonpyo and Kim, Dongju and Lu, Yan and others},
  year = 2019,
  month = {June},
  pages = {3967--3976}
}

@inproceedings{li2020,
  title = {Local Correlation Consistency for Knowledge Distillation},
  booktitle = {Proc. ECCV},
  author = {Li, Xiaojie and Wu, Jianlong and Fang, Hongyu and others},
  year = 2020,
  pages = {18--33}
}

@article{gou2025,
  title = {Neighborhood Relation-Based Knowledge Distillation for Image Classification},
  author = {Gou, Jianping and Xin, Xiaomeng and Yu, Baosheng and others},
  year = 2025,
  journal = {Neural Networks},
  volume = {188},
  pages = {107429}
}

@InProceedings{Chen_2021_CVPR,
    author    = {Chen, Pengguang and Liu, Shu and Zhao, Hengshuang and Jia, Jiaya},
    title     = {Distilling Knowledge via Knowledge Review},
    booktitle = {Proc. CVPR},
    year      = {2021},
    pages     = {5008-5017}
}

@inproceedings{gu2022,
  title = {Open-vocabulary Object Detection via Vision and Language Knowledge Distillation},
  author = {Gu, Xiuye and Tsung-Yi Lin and Weicheng Kuo and others},
  booktitle = {Proc. ICLR},
  year = 2022
}

@inproceedings{pmlr-v139-radford21a,
  title = {Learning Transferable Visual Models From Natural Language Supervision},
  author = {Radford, Alec and Kim, Jong Wook and Hallacy, Chris and others},
  booktitle = {Proc. ICML},
  volume = 	 {139},
  pages = {8748--8763},
  year = 2021
}

@inproceedings{pmlr-v139-jia21b,
  title = {Scaling Up Visual and Vision-Language Representation Learning With Noisy Text Supervision},
  author = {Jia, Chao and Yang, Yinfei and Xia, Ye and others},
  booktitle = {Proc. ICML},
  volume = 	 {139},
  pages = {4904--4916},
  year = 2021
}

@inproceedings{Wu_2023_ICCV,
  author = {Wu, Kan and Peng, Houwen and Zhou, Zhenghong and others},
  title = {{TinyCLIP}: CLIP Distillation via Affinity Mimicking and Weight Inheritance},
  booktitle = {Proc. ICCV},
  year = 2023,
  pages = {21970-21980}
}

@inproceedings{He_2016_CVPR,
  author = {He, Kaiming and Zhang, Xiangyu and Ren, Shaoqing and others},
  title = {Deep Residual Learning for Image Recognition},
  booktitle = {Proc. CVPR},
  year = 2016
}

@inproceedings{Sandler_2018_CVPR,
  author = {Sandler, Mark and Howard, Andrew and Zhu, Menglong and others},
  title = {MobileNetV2: Inverted Residuals and Linear Bottlenecks},
  booktitle = {Proc. CVPR},
  year = 2018
}

@misc{hinton2015distillingknowledgeneuralnetwork,
  title = {Distilling the Knowledge in a Neural Network}, 
  author = {Hinton, Geoffrey and Vinyals, Oriol and Dean, Jeff},
  year = 2015,
  note = {arXiv:1503.02531}
}

@misc{romero2015fitnetshintsdeepnets,
  title = {{FitNets}: Hints for Thin Deep Nets}, 
  author = {Romero, Adriana and Ballas, Nicolas and Kahou, Samira Ebrahimi and others},
  year = 2015,
  note = {arXiv:1412.6550}
}

@inproceedings{Khattak_2023_CVPR,
  author = {Khattak, Muhammad Uzair and Rasheed, Hanoona and Maaz, Muhammad and others},
  title = {{MaPLe}: Multi-Modal Prompt Learning},
  booktitle = {Proc. CVPR},
  year = 2023,
  pages = {19113-19122}
}

@inproceedings{Li_2024_CVPR,
  author = {Li, Feng and Jiang, Qing and Zhang, Hao and others},
  title = {Visual In-Context Prompting},
  booktitle = {Proc. CVPR},
  year = 2024,
  pages = {12861-12871}
}

@techreport{wah_branson_welinder_perona_belongie_2011,
  title       = {{The Caltech-UCSD Birds-200-2011 Dataset}},
  author      = {Wah, Catherine and Branson, Steve and Welinder, Peter and Perona, Pietro and Belongie, Serge},
  year        = {2011},
  institution = {California Institute of Technology},
  number      = {CNS-TR-2011-001},
  month       = {July}
}

@inproceedings{parkhi2012cats,
  title = {Cats and dogs},
  author = {Parkhi, Omkar M and Vedaldi, Andrea and Zisserman, Andrew and others},
  booktitle = {Proc. CVPR},
  pages = {3498--3505},
  year = 2012
}

@InProceedings{khosla2011novel,
  author = 	 "Khosla, Aditya and Jayadevaprakash, Nityananda and Yao, Bangpeng and others",
  title = 	 "Novel dataset for fine-grained image categorization: Stanford dogs",
  booktitle = 	 "Proc. CVPR Workshop on Fine-Grained Visual Categorization (FGVC)",
  year = 	 "2011",
  volume = 	 "2",
  number = 	 "1"
}

@article{maji2013fine,
  title = {Fine-grained visual classification of aircraft},
  author = {Maji, Subhransu and Rahtu, Esa and Kannala, Juho and others},
  journal = {arXiv:1306.5151},
  year = 2013
}

@inproceedings{deng2009imagenet,
  title = {Imagenet: A large-scale hierarchical image database},
  author = {Deng, Jia and Dong, Wei and Socher, Richard and others},
  booktitle = {Proc. CVPR},
  pages = {248--255},
  year = 2009
}

@inproceedings{liu2022convnet,
  title = {A convnet for the 2020s},
  author = {Liu, Zhuang and Mao, Hanzi and Wu, Chao-Yuan and others},
  booktitle = {Proc. CVPR},
  pages = {11976--11986},
  year = 2022
}

@inproceedings{loshchilov2018decoupled,
  title = {Decoupled Weight Decay Regularization},
  author = {Loshchilov, Ilya and Hutter, Frank},
  booktitle = {Proc. ICLR},
  year = 2019
}

@inproceedings{loshchilov2016sgdr,
  title = {{SGDR}: Stochastic Gradient Descent with Warm Restarts},
  author = {Ilya Loshchilov and Frank Hutter},
  booktitle = {Proc. ICLR},
  year = 2017
}

@article{vandermaaten2008visualizing,
  title = {Visualizing data using t-{SNE}},
  author = {Van der Maaten, Laurens and Hinton, Geoffrey},
  journal = {Journal of Machine Learning Research},
  volume = {9},
  year = 2008,
  pages = "2579--2605"
}

@InProceedings{Huang_2023_ICCV,
    author    = {Huang, Zeyi and others},
    title     = {A Sentence Speaks a Thousand Images: Domain Generalization through Distilling CLIP with Language Guidance},
    booktitle = {Proc. ICCV},
    year      = {2023},
    pages     = {11685--11695}
}

\end{document}